\documentclass[runningheads]{llncs}
\usepackage{comment}
\usepackage{multirow,hhline}
\usepackage{graphicx}
\usepackage{amsmath,amsfonts,amssymb,bm}
\usepackage{subfigure}
\usepackage{cleveref}

\begin{document}
\title{A Novel Convolutional Neural Network Architecture with a Continuous Symmetry}

\def\CICAISubNumber{29}  

\titlerunning{A Novel ConvNet Architecture with a Continuous Symmetry}
%
\author{Yao Liu\inst{1} \and
Hang Shao\inst{2} \and
Bing Bai\inst{1}}
\authorrunning{Y. Liu \emph{et al.}}
%
\institute{Tsinghua University \and Zhejiang Future Technology Institute (Jiaxing) \\
\email{liuyao@gmail.com shaohang@zfti.org.cn baibing12321@163.com}} 
\maketitle              

\begin{abstract}
This paper introduces a new Convolutional Neural Network (ConvNet) architecture inspired by a class of partial differential equations (PDEs) called quasi-linear hyperbolic systems. With comparable performance on the image classification task, it allows for the modification of the weights via a continuous group of symmetry. This is a significant shift from traditional models where the architecture and weights are essentially fixed. We wish to promote the (internal) symmetry as a new desirable property for a neural network, and to draw attention to the PDE perspective in analyzing and interpreting ConvNets in the broader Deep Learning community.

\end{abstract}
\section{Introduction}

With the tremendous success of Deep Learning in diverse fields from computer vision~\cite{he2016deep} to natural language processing~\cite{vaswani2017attention}, the model invariably acts as a black box of numerical computation~\cite{bai2021attentions}, with the architecture and the weights largely fixed, i.e., they can not be modified without changing the output (for a fixed input), except by permuting neurons or units from the same layer. One would say that the {\bf symmetry} of the neural network, the set of transformations that do not affect the model's prediction on any input, is 
$$\operatorname{Sym}(\mathrm{model}) = \prod_{i} S_{n_i}\,,$$
the product of symmetric groups on $n_i$ ``letters,'' where $n_i$ is the number of \emph{interchangeable} neurons in the $i$-th layer.

Although quite large as a group, it does not permit us to modify the model in any meaningful way. In the case of Convolutional Neural Networks (ConvNets), the channels are essentially fixed and frozen in place, due to the presence of coordinate-wise activation functions~\cite{zhang2018efficient} (such as ReLU), which arguably build certain semantic contents in the channels~\cite{olah2020zoom}, thus mixing them would destroy the model. The \emph{nonlinear} activation unit is generally thought to be an essential component for the neural network to fit arbitrary \emph{nonlinear} functions.

\begin{figure}[htb]
    \centering
        \includegraphics[width=0.9\linewidth]{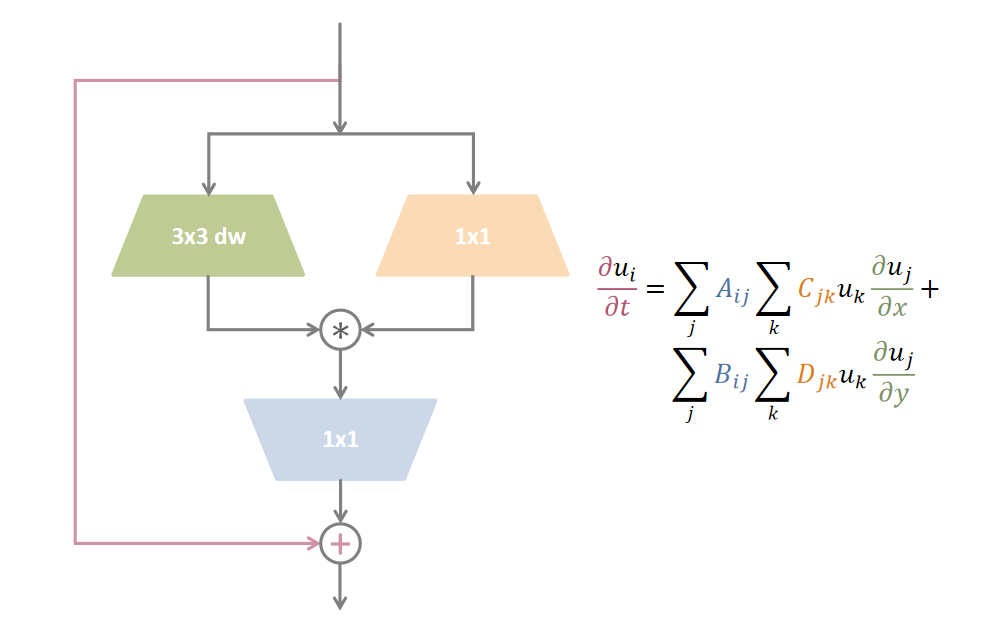}
    \caption{Schematic of a single block of our ConvNet architecture based on Eq.~(3), to replace the bottleneck block of ResNet50. The trapezoidal shapes represent the increase/decrease in the number of channels. The corresponding components of the equation are color-coded.}
\end{figure}

With inspiration and guidance from partial differential equations (PDEs), specifically \emph{first-order quasi-linear hyperbolic systems}~\cite{alinhac2009hyperbolic}, we introduce a new architecture of ConvNet with a different type of nonlinearity, which allows us to remove the activation functions without degrading performance. As a result, the new architecture admits a \emph{continuous} group of symmetry (i.e., a Lie group, as opposed to a discrete group) that allows mixing of the channels; in one version, it is the full \emph{general linear} (GL) \emph{group}, the set of all invertible $n_i \times n_i$ matrices:
$$ \operatorname{Sym}(\mathrm{model}) = \prod_{i} GL(n_i, \mathbb R)\,.$$
With a judicious choice of such transformations, one may alter the weights so that the connections become more sparse, resulting in a smaller model (a kind of lossless pruning). Since the group is continuous, one might use the method of gradient descent to search for it. In addition, it may also lead to a better understanding of the inner workings of the neural network, much like how matrix diagonalization leads to the decoupling of a system of (linear) differential equations into different ``modes'', which are easier to interpret.

We primarily present the simplest version of our model based on ResNet50, illustrated in Fig.~1 alongside the corresponding PDE. The nonlinearity is at the element-wise multiplication of the two branches (3x3 and 1x1 convs), and we apply activation functions \emph{only} at the end of 4 of the 16 blocks. See \S 5.1 for details.

This is a preliminary report on our new architecture, and the relevant parts of partial differential equations behind it. It is our hope that the research community builds upon and analyzes the properties of this new architecture, and to take the PDE perspective seriously in designing, analyzing, or simply describing \emph{all} aspects of ConvNets. Moreover, given that the Transformer architecture~\cite{vaswani2017attention} also involves a similar kind of nonlinearity apart from softmax and activation functions, it could potentially be made to admit a continuous symmetry as well.

\section{Related Work}

The link with \textbf{differential equations} has been recognized, and well exploited~\cite{weinan2017proposal,long2018pde,ruthotto2020deep,chen2018neural,weinan2022dynamical}, soon after the introduction of ResNet~\cite{he2016deep} by K. He \emph{et al.} in 2015, if not known implicitly before; see also \cite{sander2022residual} for a more recent analysis. With a few exceptions \cite{long2018pde,ruthotto2020deep}, most discussions do not make an emphasis on \emph{partial} differential equations, and to the best of our knowledge, little activity has been devoted to \emph{designing} new architecture from this perspective. Even though a PDE can be regarded as an ODE in which the state space is ``infinite-dimensional'', or upon discretization, a finite but large system with interactions only between neighboring ``pixels,'' we find the PDE perspective, specifically of hyperbolic systems, more illuminating and \emph{fruitful} (albeit limited to ConvNets), and deserves more attention and further study in the broader Deep Learning community.

A related but distinct field of research is using Deep Learning methods to solve various PDEs of interests to physicists, applied mathematicians, and engineers. We shall only mention two pioneering works that have attracted the most attention: Fourier Neural Operators (FNO) \cite{li2020fourier} and Physics-informed Neural Networks (PINN) \cite{raissi2019physics}.

\textbf{Symmetry} and \textbf{equivariance} often appear in the theoretical discussions of ConvNets and Graph Neural Networks (GNN)~\cite{cohen2016group,bronstein2021geometric}, though it is worth pointing out the distinction from our usage: More often, one says that a neural network has a (translational or permutation) \emph{symmetry}, or is \emph{equivariant} or \emph{invariant} (under translations or permutations), if when we transform the input in a certain way, the output is also transformed accordingly, or does not change at all; the model itself remains fixed. In our scenario, we are directly transforming the weights of the model, which incidentally does not need to be trained. Nevertheless, much of our work involves finding a good enough model that achieves comparable performance on standard training sets. (It shall be apparent that, as with conventional ConvNets, our model is also equivariant under translations.)

One type of operation, known under the term \textbf{structural reparametrization} \cite{ding2021diverse}, can claim to modify the model after training. However, it can only merge consecutive layers or operations that are both linear; the basic example is conv followed by batchnorm. As such, it is better regarded as a trick in training: for whatever reason, it is better to train with a deeper and more complicated architecture than is necessary for the model, and is fundamentally different from the kind of symmetry that our model has. 

\section{Designing ConvNets from the PDE perspective}

Given that ResNet is numerically solving a particular system of PDEs,
\begin{align*}
\frac{\partial u_i}{\partial t} \, =& \, \sigma\biggl(\sum_{j}L_{ij} u_j\biggr)\,, \qquad\qquad i=1,\ldots, n \\
L_{ij} \, :=& \, \alpha_{ij}\frac{\partial}{\partial x}+\beta_{ij}\frac{\partial}{\partial y} +\gamma_{ij}\frac{\partial^2}{\partial x \partial y}  + \cdots
\end{align*} 
of $n$ unknowns $u_i\equiv u_i(x,y,t)$, with initial condition at $t = 0$, in which the coefficients $\alpha_{ij}, \ldots $ are the parameters to be learned (for background, see Appendix), it is natural to take inspiration from other PDEs as found in mathematics and physics, and to see what new ConvNet architecture would come out. Here are some natural changes that one could make:

\begin{itemize}
\item[$\bullet$] Make the coefficients \emph{variables} (of $x$ and $y$), e.g., simply as linear or polynomial functions. The equation would still be linear, but now the space of PDEs would include this special equation: ($n=1$)
$$ \frac{\partial u}{\partial t} = -y\, \frac{\partial u}{\partial x} + x\, \frac{\partial u}{\partial y} \,, $$
which is solved by simply \emph{rotating} the initial data $f(x,y)$ by angle $t$ (around the origin):
$$u(x,y,t) = f(x\cos t - y \sin t, x \sin t + y\cos t) \,,$$
as can be readily verified. It is reasonable to expect that such a variation on ResNet would allow the model to make \emph{rotations} and \emph{dilations} --- in addition to translations --- on the input image.

\item[$\bullet$] The standard ``zero padding'' of conv layer seems to correspond to the so-called \emph{Dirichlet} boundary condition: the values of $u$ on the boundary are prescribed for all $t>0$. On closer inspection, it is slightly different. Furthermore, one could also experiment with other boundary conditions; the other natural one in physics is the \emph{Neumann} condition, that the \emph{normal derivative} of $u$ on the boundary is prescribed. The different conditions have the effect that the signals would ``bounce back'' off the boundary differently.

\item[$\bullet$] In a typical PDE, the matrix of coefficients is constant or slowly varying with time $t$, while in neural networks the weights from different layers are initialized independently, drawn from a (normal) distribution. One could try to force the weights from neighboring layers to correlate, either by weight-sharing or by introducing a term in the loss function that penalizes large variations between layers.

\end{itemize}

Having experimented with some of these ideas on small datasets, we did not find a specific variation that yields convincing results on the full ImageNet. We then looked into ways that the coefficients may depend on $u$ itself, which makes the equation nonlinear (apart from the activation functions). It may be viewed as a kind of ``dynamic kernel,'' but we draw inspiration from a class of PDEs called \emph{quasi-linear hyperbolic systems}, which may be the simplest, well-studied nonlinear systems in which the number of equations can be arbitrary.

In two spatial and one time dimensions, a first-order {\bf quasi-linear system} is (typically) of the form
\begin{equation}
    \frac{\partial u_i}{\partial t} = \sum_j \mathcal A_{ij}(u)\frac{\partial u_j}{\partial x} + \sum_j \mathcal B_{ij}(u) \frac{\partial u_j}{\partial y} \,,
\end{equation}
where the coefficient matrices may depend on $u$ (but not derivatives of $u$), and it is {\bf hyperbolic}\footnote{The designation may sound cryptic. It originates from the classic wave equation, which has some semblance in form with the equation of a hyperbola or hyperboloid; see Appendix \S A.3. To avoid any confusion, it is \emph{not} related to hyperbolic plane/space/geometry that has made into machine learning} if any linear combination of $\mathcal A$ and $\mathcal B$ is diagonalizable with only \emph{real} eigenvalues (e.g., $\mathcal A$ and $\mathcal B$ are symmetric) for \emph{any} value of $u$. Leaving aside the latter condition, the simplest example is to make each entry a linear function of $u$: 
\begin{equation}
    \mathcal A_{ij}(u) = \sum_{k} \mathcal A_{ijk} u_k \,,
\end{equation}
and similarly for $\mathcal B$. By dimension count, such a tensor would be very large (for large $n$), and it would deviate too much from typical ConvNets. Instead, we shall restrict to 
\begin{equation} \frac{\partial u_i}{\partial t} = \sum_{j} A_{ij} \sum_{k} C_{jk} u_k \frac{\partial u_j}{\partial x} + \sum_{j} B_{ij} \sum_{k} D_{jk} u_k \frac{\partial u_j}{\partial y} \,, 
\end{equation}
which is straightforward to turn into a ConvNet (see \S 5.1 and Fig.~1 for details), and the number of parameters is kept at a reasonable level. Since nonlinearity is already built-in, we thought it would not be necessary to add activation functions, at least not at every turn; and much to our surprise, the model trains just as well, if not better. With this simple change, the model now has a \emph{continuous} symmetry, from mixing of the channels, that is not present in conventional ConvNets with coordinate-wise activation functions at every conv layer, and we believe it is a more significant contribution than matching or breaking the state of the art (SOTA). It is likely that, with enough compute, ingenuity, and perhaps techniques from Neural Architecture Search, variations of this architecture could compete with the best image models of comparable size. (We have not tried to incorporate the modifications listed earlier, as they are all linear, and we wish to explore the new nonlinearity on its own.)

It is observed that, once we remove all the activation functions, or use the standard ReLU, the training is prone to breaking down: it would fail at a particular epoch, with one or more samples (either from the training or validation sets) causing the network to output \texttt{NaN}, and the model could not recover from it. To mitigate this, we add activation functions such as hardtanh that clip off large values, only once every few blocks. It is also observed that resuming training with a smaller learning rate may get around the ``bad regions'' of the parameter space. More analyses are needed to determine the precise nature and cause of this phenomenon (it might be related to the formation of ``shock waves'' in nonlinear hyperbolic equations \cite{alinhac2009hyperbolic}), and perhaps other ways to circumvent it.

We provide here the details of the activation functions that we experimented with. In standard PyTorch \cite{paszke2019pytorch}, \texttt{nn.Hardtanh} is implemented as
$$ 
    \operatorname{hardtanh}(x) := \begin{cases}
     \mathrm{max\_val} & \text{if } x > \mathrm{max\_val} \\
     \mathrm{min\_val} & \text{if } x < \mathrm{min\_val} \\
     x & \mathrm{otherwise} \,.
     \end{cases}
$$
We typically use $\pm 1$ as the clip-off values. We also introduce two multi-dimensional variants that we call ``hardball'' and ``softball,'' which appear to give better performance. Hardball is so defined that it takes a vector $\mathbf{x} \in \mathbb R^n$ and maps it into the ball of radius $R$,
$$ \operatorname{hardball}(\mathbf{x}) := \begin{cases}
    \mathbf{x} & \text{if } |\mathbf{x}| < R \\
    R \mathbf{x} / |\mathbf{x}| & \text{if } |\mathbf{x}| \geq R \,,
\end{cases} $$
where $|\mathbf{x}|$ is the Euclidean norm. We set $R$ to be the square root of $n$ (the number of channels), though other choices may be better. Softball is a soft version,
$$ \operatorname{softball}(\mathbf{x}) := \frac{\mathbf{x}}{\sqrt{1+|\mathbf{x}|^2/R^2}} \,, $$
and they are both spherically symmetric. (One may perhaps regard them as normalization layers rather than activation functions.)

\section{Symmetry of the model}

The symmetry of our model would depend on the specific implementation, and may be more complicated than one would naively expect. We shall first consider it on the level of the PDE.

With a change of coordinates $\tilde u_i = \sum_j T_{ij} u_j$ for an invertible matrix $T$, the general equation (1) with (2)
$$\frac{\partial u_i}{\partial t} = \sum_{j,k} \mathcal A_{ijk} u_k \frac{\partial u_j}{\partial x} + \sum_{j,k} \mathcal B_{ijk} u_k \frac{\partial u_j}{\partial y} $$
 would transform \emph{only} in the coefficient tensors $\mathcal A_{ijk}$ and $\mathcal B_{ijk}$. Indeed, 
 
 \noindent (For clarity, we omit the second half involving $\mathcal B$ and $\frac{\partial}{\partial y}$.)
\begin{align*}
    \frac{\partial \tilde u_i}{\partial t} &= 
        \sum_{j}T_{ij} \frac{\partial u_j}{\partial t} \\
    &= \sum_{j}T_{ij} \sum_{k,l} \mathcal A_{jkl} u_l \frac{\partial u_k}{\partial x} \\
    &= \sum_{j}T_{ij} \sum_{k,l} \mathcal A_{jkl} \sum_{m} T^{-1}_{lm} \tilde u_m \sum_r T^{-1}_{kr}  \frac{\partial \tilde u_r}{\partial x}\\
    &= \sum_{m,r} \Biggl( \underbrace{ \sum_{j,k,l} T_{ij} \mathcal A_{jkl} T^{-1}_{lm} T^{-1}_{kr}  }_{ \let\scriptstyle\textstyle \substack{\tilde{\mathcal A}_{irm}}} \Biggr) \tilde u_m \frac{\partial \tilde u_r}{\partial x} \,.
\end{align*}
We note in passing that similar calculations are commonplace in \emph{classical} differential geometry when making a change of coordinates \emph{on the base space}; here, we are making a change of coordinates on the ``dependent'' variables. From a more abstract point of view, this is the induced representation of $GL(V)$ on the tensor product $V\otimes V^* \otimes V^* \cong \operatorname{Hom}(V\otimes V, \, V)$ for a vector space $V\cong\mathbb R^n$.

On the level of the neural network, we only need to make sure that the $T^{-1}$ comes from the previous layer, i.e., it is the inverse of the $T$ that appears in transforming the previous block (if different). With such a transformation at each block, we find that the overall symmetry of the model is
$$ \operatorname{Sym}(\text{model}) = \prod_i G_i \,,$$
with
\begin{equation*} 
    G_i = 
    \begin{cases}
        S_{n} & \sigma=\text{relu, or any element-wise activation function} \\
        O(n) & \sigma=\text{hardball, softball, etc.\footnotemark} \\
        GL(n, \mathbb R) & \sigma=\text{identity} \,.
    \end{cases}
\end{equation*}
\footnotetext{The group $O(n):=\{ M \in GL(n, \mathbb R) \;|\; MM^T = I_n\}$ is known as the \emph{orthogonal group}, and it preserves the Euclidean norm on $\mathbb R^n$.}

As noted before, this ``fully connected'' block would be too costly to train, if we are to match ResNet in which the last stage uses as many as $n=512$ channels. One simple way to reduce the number of parameters is to make the tensor ``block diagonal'', and the transformation would only mix channels from the same block. The bottleneck block of ResNet achieves this by shrinking the number of channels before applying the 3x3 conv, but a similar approach would introduce additional layers that shield the main operations that we would like to transform.

If we are to take $\mathcal A_{ijk}$ to be of the special form as in Eq.~(3), i.e., as the product of two matrices $A_{ij}C_{jk}$ (no summation), then it is not guaranteed that the transformed tensor would still factorize in the same way, for a generic $T$. By simple dimension count, the set of tensors that are factorizable in the prescribed way is a subvariety (submanifold) of dimension at most $2n^2$, and one wishes to find a $T\in GL(n,\mathbb R)$ that keeps the resulting $\tilde{\mathcal A}_{ijk}$ on the subvariety. Given that the dimension of $GL(n,\mathbb R)$ is $n^2$ and that $2n^2 + n^2 \ll n^3$, it is not \emph{a priori} obvious that such transformations exist, apart from simple scalings and $S_n$, that are universal for all $\mathcal A_{ijk}$.

What one may hope for is that, for a specific $\mathcal A_{ijk}$ that factors, we can find such a $T$. For example, if for some pair of indices $j,j'$, we have $A_{ij}C_{jk}=A_{ij'}C_{j'k}$ for all $i,k$, then we can perform a rotation in the plane of the $j$ and $j'$ directions:
\begin{equation*}
\begin{cases}
 \tilde u_j = \alpha u_j + \beta u_{j'} \\
 \tilde u_{j'} = \gamma u_j + \delta u_{j'}
\end{cases}
\qquad \alpha\delta - \beta\gamma \neq 0 \,.
\end{equation*}
Further investigation, either theoretical or numerical, may be needed to answer this question satisfactorily. It may be the case that there exists a symmetry that preserves the output \emph{not} for all inputs, but only those inputs that are ``similar'' to the dataset (``in distribution''). It would be a weaker form of symmetry, but no less useful in practice.

Lastly, it should be remarked that there is a trivial ``symmetry'' in the tensor $\mathcal A_{ijk}$ in the last two indices, i.e., $\mathcal A_{ijk}$ and $\mathcal A_{ikj}$ can be interchanged (so long as their sum is fixed). One may regard this as a redundancy in the parameter space, for we can force $\mathcal A_{ijk}=\mathcal A_{ikj}$, or $\mathcal A_{ijk}=0$ for $j<k$ (and reduce the dimension roughly by half), and not due to mixing of the channels. We have not exploited this in the present work.

\section{Experimental Results}

\subsection{Details of the Architecture}

How do we turn Eq.~(3) into a ConvNet? We first make the differential operators $\partial/\partial x$ and $\partial/\partial y$ into 3x3 convolutional kernels, but we allow the weights to be trainable instead of fixed. Each incoming channel (each $u_j$) would split into 2 --- or better, 4 --- channels, and this is conveniently implemented in \texttt{nn.Conv2d} by setting \texttt{groups} to equal the number of input channels, as in ``depthwise convolution'' \cite{chollet2017xception}. The matrices are simply 1x1 conv layers, with $A$ and $B$ stacked into one, and $C$ and $D$ stacked into one. Batchnorm is applied after the 3x3 and at the end. As with standard ResNet, the time derivative turns into the skip connection, and we arrive at the architecture of a single block as illustrated in Fig.~1. We do not enforce symmetry of these matrices to make the equation hyperbolic in the technical sense. As a general rule, we need not strictly follow the equation, but take the liberty in relaxing the weights whenever convenient.

One novelty is to make the weights of the 3x3 conv shared across the groups, which would make the claimed symmetry easier to implement. One may achieve this by making an \texttt{nn.Conv2d} with 1 input channel and 4 output channels, and at forward pass, we ``repeat'' the $4\times 1\times 3\times 3$ weights in the zeroth dimension before acting on the input tensor. Most of our experiments are with this ``minimalist'' 3x3 conv, except the ones marked ``no ws'' (no weight-sharing) in Table~1.

It may be possible to implement this kind of ``variable-coefficient convolution'' \emph{natively}, instead of using the existing \texttt{torch.nn} layer which is tailored to conventional convolutions.

For the full design of the neural network, we simply take the classic ResNet50 with \texttt{[3,4,6,3]} as the numbers of blocks in the four stages. No activation function is applied except once only in each stage (e.g., at each downsampling), and we use \texttt{nn.Hardtanh} or our variants, hardball and softball (see \S 3 for definitions), instead of ReLU, lest the training would fail completely or the resulting model would not work as well. 

\subsection{Experiments}

As is standard in computer vision since the 2012 Deep Learning revolution, we have trained our model as an image classification task on the ImageNet dataset. The ImageNet-1k contains 1000 classes of labeled images, and for the sake of faster iterations, we primarily trained on a 100-class subset on a single GPU, while maintaining the standard image size of 224x224.

We use the \texttt{timm} library of PyTorch image models \cite{rw2019timm} for best practices in implementing ResNet and its training \cite{wightman2021resnet}. We took the official training script for ResNeXt-50 (SGD, cosine learning rate, with a warmup of 5 epochs, batch size of 192, etc.) except that the peak learning rate is set to 0.3 instead of 0.6, and the total number of epochs is set to 50. The results are in Table~1, where we mainly record two ways of modifying the model: changing only the activation function, and altering the placements of the conv layers within the block, corresponding to modifying Eq.~(3) into Eqs.~(4)--(7):

\begin{align} 
    \frac{\partial u_i}{\partial t} &= \sum_{k} C_{ik} u_k \sum_{j} A_{ij} \frac{\partial u_j}{\partial x} + \sum_{k} D_{ik} u_k \sum_{j} B_{ij} \frac{\partial u_j}{\partial y} \\
    \frac{\partial u_i}{\partial t} &= \sum_{k} C_{ik} u_k  \frac{\partial u_i}{\partial x} +  \sum_{k} D_{ik} u_k \frac{\partial u_i}{\partial y} \\  
    \frac{\partial u_i}{\partial t} &= \sum_{j} A_{ij} \frac{\partial }{\partial x} \sum_{k} C_{jk} u_j u_k  + \sum_{j} B_{ij} \frac{\partial }{\partial y}  \sum_{k} C_{jk} u_j u_k \\  
    \frac{\partial u_i}{\partial t} &= \sum_{j} A_{ij} \frac{\partial }{\partial x} \sum_{k} C_{jk} u_j u_k  + \sum_{j} B_{ij} \frac{\partial }{\partial y}  \sum_{k} D_{jk} u_j u_k 
\end{align}

\begin{table}[ht]
\caption{Performance on a 100-class subset of ImageNet-1k, trained for 50 epochs with identical training strategy. For our model, activation is applied either at the end of each block (@all), or only at downsampling (@ds). Inside the block, the number of channels increases by a factor of 4, except when indicated with ``x6''.}
\label{tab1}
\begin{center}
\begin{tabular}{cccc}
     model &  \#parameters\phantom{ } &  top1-acc & activation \\
\hline
    ResNet50 & 23.7M & 84.52 & \\
    MobileNet\_v3\_large & 4.33M & 82.91 &  \\
\hline
    Eq.~(3) & 8.61M & 82.06 & relu@all \\
    Eq.~(3) & 8.61M & 82.34 & hardtanh@ds \\
    Eq.~(3) & 8.61M & 83.50 & hardball@ds \\
    Eq.~(3) & 8.61M & 83.66 & softball@ds \\
    Eq.~(3) (no ws) & 8.73M & 84.24 & softball@ds \\
    Eq.~(3) (no ws, x6) & 13.0M & 84.58 & softball@ds \\
\hline
    Eq.~(4) & 5.70M & 81.88 & \multirow{4}*{softball@ds} \\
    Eq.~(5) & 4.26M & 78.64 &  \\
    Eq.~(6) & 5.61M & 82.52 &  \\
    Eq.~(7) (no ws, x6) & 13.0M & \bf 84.96 &  \\
\end{tabular}
\end{center}
\end{table}

Note that Eq.~(5) only has each $u_i$ depending on its own derivatives, and the model is thus smaller and limited in expressivity or capacity. Eqs.~(6) and (7) have the derivative acting on the product $u_j u_k$, and it is often called a system of \emph{conservation laws}. They can easily be rewritten in the form of Eq.~(3), and the difference in performance may be attributable simply to the model size.

It is expected that, when going to the full ImageNet-1k, and allowing for longer training and hyperparameter tuning, the best-performing model may be different from the ones we found in Table~1. 

We refrain from making assertions on \emph{why} --- or \emph{if} --- this class of PDEs is superior as a base model for ConvNets, or \emph{how} the theory of PDE can provide the ultimate answer to the \emph{effectiveness} of neural networks as universal function approximators by means of gradient descent. Whatever mechanisms that make ResNet work, also make our model work.

\section{Conclusion}

We present a new ConvNet architecture inspired by a class of PDEs called quasi-linear hyperbolic systems, and with preliminary experiments, we found a simple implementation that showed promising results. Even though it is known, within small circles, the close connection between PDE and ConvNet, we made the first architecture design directly based on a nonlinear PDE, and as a result, we are able to remove most of the activation functions which are generally regarded as indispensable. The new architecture admits a continuous symmetry that could be exploited, hopefully in future works. We expect that this work opens up a new direction in neural architectural design, demonstrates the power of the PDE perspective for ConvNets, and opens the door for other concepts and techniques in nonlinear PDE, both theoretical and numerical, for improved understanding of neural networks.

\section*{Acknowledgements}
This work is supported in part by Provincial Key R\&D Program of Zhejiang under contract No.~2021C01016, in part by Young Elite Scientists Sponsorship Program by CAST under contract No.~2022QNRC001.

\section*{Addendum}

It has come to our attention that the recent work \cite{zhao2023symmetries}, \&c., are highly relevant though largely complementary to ours, in terms of exploiting the kind of continuous symmetry presented here. Specifically, our activation functions, hardball and softball, are examples of what they call \emph{radial} activation functions, and the symmetries are more aptly referred to as \emph{symmetries in the parameter space}. 

Not surprisingly, the operation of element-wise multiplying two branches of the input had been proposed independently by several authors: polynomial networks or $\Pi$-Nets \cite{chrysos2020p,chrysos2021deep} in which the output is a polynomial function of the input, PDO-eConvs \cite{he2022neural} that comes from the equivariance perspective, and what is called the \emph{gated linear unit} (GLU) \cite{dauphin2017language,shazeer2020glu} that has made its appearance in recent language models, though not without antecedents in the winters of neural networks. We apologize for our omissions in the published version.


\bibliographystyle{splncs04}
\bibliography{mybibliography}

\clearpage
\appendix

\section{Appendix: A Crash Course on PDE}

We shall give a short, gentle, yet (mostly) self-contained primer on the elementary facts of PDEs that are most relevant to Deep Learning. There is no original content, except for the overall scope, choice of presentation, and certain points which may not be easy to find in one place. The reader who is familiar may skip \S A.1, with a glance at \Cref{tab2}. The mathematical field of PDE is staggeringly vast, and any account is by necessity limited, sometimes with the same term (e.g., hyperbolicity) appearing to have very different definitions in different accounts. As with any area of mathematics, it may appear to the uninitiated more complicated than it actually is, with the level of generality and confusing (even conflicting) notations that it operates in. Luckily, we shall not need any of the heavy technical machineries; the key is to form the bridge, and lots of engineering and experimentation with training neural networks.

We shall take advantage of a small simplification: we will exclusively be dealing with \emph{two} spatial dimensions ($x$ and $y$), for a ConvNet takes a 2D image as input, and each subsequent layer (before the penultimate layer) has a 2D structure to it.

\subsection{First example: the heat equation}

It is instructive, perhaps even necessary, to begin with a consideration of the \emph{heat equation} on, say, a two-dimensional metal plate:
$$\frac{\partial u}{\partial t} = \mathrm{\Delta} u \,.$$
Here $u$ is the temperature at a point in space \emph{and} time, i.e., $u\equiv u(x,y,t)$, and $\mathrm{\Delta}$ is the Laplacian operator:
$$\mathrm{\Delta} u := \frac{\partial^2 u}{\partial x^2} +  \frac{\partial^2 u}{\partial y^2} \,.$$
The $x$ and $y$ coordinates extend over the domain, say from $[0,1]$ for a square-shaped plate, and $t$ goes from $0$ to $\infty$. The equation of the unknown function $u$ is called a \emph{partial differential equation}, since it involves the partial derivatives of $u$; and by a \emph{solution} we simply mean a specific function $u$ that \emph{solves} or satisfies the equation, for all values of $x, y$, and $t$.

The heat equation describes (or models) how temperature evolves: if you are at the bottom of a valley, your temperature should go up, and if you are at the top of a hill, your temperature should come down. It is found that, up to some proportionality constant, this captures well how heat diffuses in the real world. 

As with \emph{any} partial differential equation, one would also need to impose some \emph{initial} and/or \emph{boundary conditions}. In this case, we specify the temperature over the whole plate at time $t=0$, and at each point on the boundary for all $t>0$; solving the equation \emph{along with} the appropriate initial (and boundary) conditions is often called the Cauchy problem, or Initial(-Boundary) Value Problem. Our physical intuition tells us that the temperature is then determined inside for all $t>0$, and it is formalized mathematically that the Cauchy problem is \emph{well-posed} (in the sense of J. Hadamard): the \emph{existence} and \emph{uniqueness} of solution which depends \emph{continuously} on the initial and boundary data.

\begin{figure}[htb]
    \centering
    \begin{subfigure}
        \centering
        \includegraphics[width=0.175\linewidth]{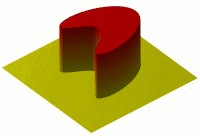}
        \hfill
        \includegraphics[width=0.175\linewidth]{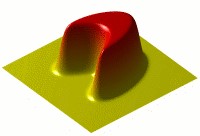}
        \hfill
        \includegraphics[width=0.175\linewidth]{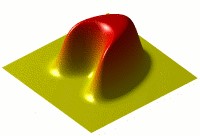}
        \hfill
        \includegraphics[width=0.175\linewidth]{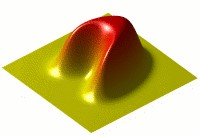}
        \hfill
        \includegraphics[width=0.175\linewidth]{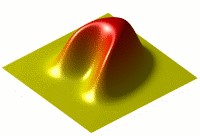}
    \end{subfigure}
    \caption{The time-lapse of a solution to the heat equation, over a square domain. The temperature is represented in the vertical direction and by the coloring. Credits: Oleg Alexandrov via Wikimedia Commons.}
\end{figure}

How does one go about solving the heat equation? There are elaborate tools for solving various, though rather limited, PDEs, starting with the Fourier series (first introduced by J. Fourier precisely to solve the heat equation); but for us it is more important to walk through the process of solving it \emph{numerically}. First, imagine we discretize the square plate into a grid, so that the temperature $u_i$ at each (discrete) time $t_i$ is represented by a 2D array of numbers. Starting with the initial data $u_0$ at $t_0=0$, we can approximate its Laplacian by
$$
\begin{aligned} 
\mathrm{\Delta}_h u_0(x,y) :=
 & \frac{u_0(x+h,y)-2u_0(x,y)+u_0(x-h,y)}{h^2} + \frac{u_0(x,y+h)-2u_0(x,y)+u_0(x,y-h)}{h^2} \\
=&  \frac{u_0(x+h, y)+u_0(x-h, y)+u_0(x, y+h)+u_0(x, y-h) - 4 u_0(x,y)}{h^2} \,,
\end{aligned} 
$$
which is used to find the temperature $u_1$ at time $t_1=\tau$:
$$ \frac{u_1 - u_0}{\tau} \approx \left.\frac{\partial u}{\partial t} \right|_{t=0} = \mathrm{\Delta} u_0 \,.$$
Iteratively, we may find the temperature $u_{i}$ at time $t_{i}$ from $u_{i-1}$,\footnote{Note that the subscript in $u_i$ throughout the Appendix represents the ``time slice'' at $t=t_i$, \emph{not} a component of the system of equations as in the main paper.} namely, 
$$u_{i} \approx u_{i-1} + \tau \mathrm{\Delta}_h u_{i-1} \,.$$
More sophisticated numerical schemes may be used instead of this simplest approach, which is usually called the (forward) Euler method.

One may have noticed that the whole process is very similar to a ResNet (with a single channel throughout) with a single kernel:
\begin{equation}
k =\frac{\tau}{h^2} \begin{bmatrix}
    0 & 1 & 0 \\
    1 & -4 & 1 \\
    0 & 1 & 0
\end{bmatrix} \,,
\end{equation}
and that the boundary condition is needed to compute the convolution $k * u_{i}$ along the boundary. One can hardly escape the conclusion that ResNet (i.e., the feedforward of ResNet) is simply a discretization of a PDE of the form
$$ \frac{\partial u}{\partial t} = \sigma (Lu) \,,$$
where $L$ is some (linear combination of) differential operators 
$$ L = \alpha \frac{\partial }{\partial x} + \beta  \frac{\partial }{\partial y} + \gamma \mathrm{\Delta}  + \cdots  \,, $$
and $\sigma$ is the nonlinear activation function. Indeed, $\frac{\partial}{\partial x}$ and $\frac{\partial}{\partial y}$ can be realized by the so-called \emph{Sobel operators}, and the space of differential operators of order $\leq 2$ is six-dimensional, well represented in the 9-dimensional space of kernels. If $u$ is a vector-valued function (for ResNet with multiple channels), then $L$ would be a matrix of differential operators, i.e., a square matrix where each entry takes the form above; more on systems later. The coefficients $\alpha, \beta, \ldots$ may depend on $t$, to account for the fact that the weights in a ConvNet differ from one layer to the next.

We shall summarize in the form of a dictionary, which hopefully suffices for understanding Fig.~1 and the main thesis of the paper.

\begin{table}
\caption{Dictionary of terminologies between ConvNet and PDE\protect\footnotemark.}
\label{tab2}
\begin{center}
\setlength{\tabcolsep}{5pt}
\begin{tabular}{c|c}
    \bf ConvNet & \bf PDE \\
\hline
    input & initial condition \\
    feed forward & solving the equation \emph{numerically} \\
    hidden layers/feature maps & solution at intermediate times \\
    final layer & solution at final time \\
    3x3 convolution & differential operator of order $\leq 2$ \\
    weights/parameters & coefficients \\ 
    padding & boundary condition \\
    groups & matrix is block diagonal \\
    residual connection & time derivative 
\end{tabular}
\end{center}
\end{table}

\footnotetext{Many items in the table apply more broadly, between general neural networks and ODEs or dynamical systems. There are also notable omissions from this table, such as batchnorm and max/avg pooling.}

It is unfortunate that ConvNets historically evolved mainly without much direct input from PDEs, resulting in some conflicts in terminology with common mathematical usage (notably, block and group). It is likely that Y. LeCun was well aware of the connection, yet he chose to present the first ConvNet using convolutions, familiar as \emph{filters} in computer vision and image processing, in lieu of differential operators.

Now, the obvious question is: What does the training of a neural network via gradient descent (the \emph{backpropagation} algorithm) correspond to in mathematics? For many applied fields that are governed or modeled by PDEs, one often does not have the full information of the underlying equation or the initial/boundary data, which one needs to \emph{infer} from the solutions that \emph{can} be observed. The problem of finding the equation from its solutions is generally known as the \emph{inverse problem}, and is a large field in and of itself. We shall, however, briefly turn to a somewhat different topic, with certain points of general interest.

\subsection{Optimal control and the calculus of variations}

A slight change of terminology (and a shift in goal) in certain applied fields has given rise to Control Theory and Optimal Control, where the variables $u$ are sometimes called the \emph{state variables}, and the parameters that can be tweaked (or learned) are called the \emph{control variables}. One may see the equation written with the explicit mention of the control variables $\theta$:
$$\frac{\partial u}{\partial t} = F(u, \nabla u, \ldots; x, y, t; \theta) $$
where one is supposed to optimize $\theta$ for a certain \emph{loss function}, possibly subject to additional constraints as the case may be. See \cite{li2020dynamical} for an introduction to Optimal Control theory in connection with Deep Learning.

In the case that the control variables form a continuum (that is, $\theta: S\to \mathbb R$ is a function over some \emph{continuous} space $S$) and the loss function takes the form of an integral
$$ \mathrm{Loss}(\theta) = \int_{S} \ell(\theta(s))\, ds \,,$$
it is a classical \emph{calculus of variations} problem for which one may derive the \emph{exact} equation (the Euler-Lagrange equation) that the optimal $\theta$ needs to satisfy, or there is an algorithm to approximate the minimum --- literally the method of gradient descent. The classic example is when $\ell(\theta)=|\nabla \theta|^2$: the exact solution is given by Laplace's equation $\mathrm{\Delta}\theta = 0$ on $S$ (with suitable boundary conditions), and the approximation is nothing but solving the heat equation $\frac{\partial\theta}{\partial t}=\mathrm{\Delta}\theta$ on $S$ --- one may refer to this as \emph{gradient flow}, in contrast to the discrete steps of gradient descent.

More commonly, however, the loss function actually depends on the solution $u$ of the state equation (or the solution at the final time, as in ConvNets), thus the dependence on $\theta$ is indirect or \emph{implicit}. A loss term of explicit $\theta$ dependence would correspond to the \emph{regularization} of the neural network model.

Much has been expounded on the \emph{curse of dimensionality} by experts in applied mathematics and scientific computing, and how neural networks seem to defy the curse \cite{weinan2021algorithms}. It might be helpful to take the view that, much like how calculus provides an efficient way to optimize a function over a \emph{large} number of points, the calculus of variations is a sort of simplification, at least on a psychological level, on optimizing a function over a \emph{large} number of dimensions:

\begin{table}[h!]
\caption{Finding the minimum of a function $f:X\to\mathbb R$, for various types of $X$.}
\begin{center}
\setlength{\tabcolsep}{3pt}
\begin{tabular}{c|ccc}
    & $ X=\{x_1,\ldots, x_n \}$ & $X = \mathbb R^n$ & 
    $X=\{\theta: S\to\mathbb R \}$ \\
\hline
    \multirow{2}*{$X$ is} & \multirow{2}*{a finite set} & a finite-dimensional & an infinite-dimensional \\
    & & space or manifold & vector space \\
\hline
    \multirow{2}*{exact solution} & comparison & calculus:  & calculus of variations:  \\
     & (pairwise) & $\nabla f = 0$ & $\delta f = 0$ \\
\hline
    approximation & -- & gradient descent & successive approximation \\
\end{tabular}
\end{center}
\end{table}

\subsection{The wave equation and hyperbolic systems}

It can be argued that the standard ResNet is most similar to a (first-order) hyperbolic system. Even though the concept of hyperbolic systems or PDEs may sound esoteric, some of the most famous equations in physics fall in this ``category'': Maxwell and Dirac are linear, while Einstein and Navier-Stokes are nonlinear examples that have been the driving forces behind a substantial amount of current research in \emph{both} theoretical and numerical PDE.

We shall begin with the prototype of hyperbolic equations: the \emph{wave equation}
$$\frac{\partial^2 u}{\partial t^2} = \mathrm{\Delta} u \,,$$
which models the up-and-down movement of a two-dimensional rubber sheet or membrane that is fixed along the boundary. To solve it, just imposing an initial condition on $u$ is not enough; we also need to specify $\frac{\partial u}{\partial t}$ at time $t=0$. Together, they give the ``first-order'' approximation to $u_1$ at time $t_1=\tau$. To find $u_2$, we note that
$$\frac{u_2-2u_1+u_0}{\tau^2} \approx \left.\frac{\partial^2 u}{\partial t^2}\right|_{t=t_1} = \mathrm{\Delta} u_1  \,,$$
which gives $u_2$ in terms of $u_1$ and $u_0$, and so on for all $u_i$:
$$ u_i \approx 2u_{i-1} - u_{i-2} + \tau^2 \mathrm{\Delta}_h u_{i-1} \,.$$
This would be a ResNet with two skip connections at each layer, or a kind of DenseNet \cite{huang2017densely} (of a single channel throughout). 

We can rewrite the wave equation as a \emph{system} of two equations:
$$ \left\{ \begin{aligned} \tfrac{\partial u}{\partial t} &= v \\
\tfrac{\partial v}{\partial t} &= \mathrm{\Delta} u  \end{aligned} \right. \qquad\text{or}\qquad
\tfrac{\partial }{\partial t} \begin{pmatrix} u \\ 
v \end{pmatrix} =  \begin{pmatrix} 0 & \phantom{.}1 \\ \mathrm{\Delta} & \phantom{.}0 \end{pmatrix} 
\begin{pmatrix} u \\ 
v \end{pmatrix} \,,$$
which makes clear that we need to specify both $u$ and $v=\frac{\partial u}{\partial t}$ at time $t=0$. Solving this system of PDEs numerically is nothing but a ResNet with two channels, with the nontrivial 3x3 kernel given by the same $k$ as in Eq.~(8).

It turns out that we can rewrite the wave equation in yet another way: with $ w = \big( \tfrac{\partial u}{\partial x}, \tfrac{\partial u}{\partial y}, \tfrac{\partial u}{\partial t} \big)^T$, one can easily verify that $w$ satisfies
\begin{equation*}
\frac{\partial w}{\partial t} = 
    \begin{pmatrix}
        0\phantom{.} & \phantom{.}0\phantom{.} & \phantom{.}1 \\
        0\phantom{.} & 0 & \phantom{.}0 \\
        1\phantom{.} & 0 & \phantom{.}0
    \end{pmatrix} \frac{\partial w}{\partial x} + 
    \begin{pmatrix}
        0\phantom{.} & \phantom{.}0\phantom{.} & \phantom{.}0 \\
        0\phantom{.} & 0 & \phantom{.}1 \\
        0\phantom{.} & 1 & \phantom{.}0
    \end{pmatrix} \frac{\partial w}{\partial y} \,, 
\end{equation*}
which is a system of \emph{first-order} equations (only the first-order derivatives of $w$ appear) in which the coefficient matrices are \emph{symmetric}. As such, it is an example of a \emph{first-order linear hyperbolic system}:
\begin{equation}
\frac{\partial w}{\partial t} =  A\, \frac{\partial w}{\partial x} + B\, \frac{\partial w}{\partial y} \,,  
\end{equation}
where any linear combination of $A$ and $B$ is diagonalizable with only \emph{real} eigenvalues. Maxwell's equations (in vacuum) and the Dirac equation (with zero mass) are of this form, but in \emph{three} spatial dimensions; and it is this form that leads to the generalization of \emph{quasilinear} hyperbolic systems of Eq.~(1), by allowing the matrices $A$ and $B$ to depend on $w$. It is also referred to as a system of \emph{conservation laws}, especially when written in \emph{divergence form}:
$$ \frac{\partial w}{\partial t} = \frac{\partial}{\partial x} F(w) + \frac{\partial}{\partial y} G(w) $$
for some nonlinear (vector-valued) functions $F$ and $G$, for which the Jacobian matrices $A(w)=\big(\frac{\partial F}{\partial w}\big)$ and $B(w)=\big(\frac{\partial G}{\partial w}\big)$ satisfy the diagonalization condition above, for any value of $w$. The prototypical examples are the Euler and Navier-Stokes equations in fluid mechanics.

The resemblance of the \emph{linear} case of Eq.~(9) to existing ConvNet architectures is in fact more pronounced than Table~2 suggests, even in the absence of second-order derivatives: rewriting the equation of $u$ in terms of $w$ corresponds to the ``stem'' that converts the RGB channels of the input into as many as 64 channels, and the separation of the coefficient matrices from the differential operators $\frac{\partial}{\partial x}$ and $\frac{\partial}{\partial y}$ corresponds to the so-called \emph{depthwise separable convolution}~\cite{howard2017mobilenets,chollet2017xception}. It should also be remarked that diagonalizing $A$ and $B$ would effect a change of variables on $w$, which offers a curious interpretation of the 1x1 conv layers in the standard bottleneck block of ResNet50.

\subsection{General theory of linear PDEs}

\begin{quote}
\textit{This section may be read for any $d$ of spatial dimensions, and we shall employ $\vec{x}$ and $\vec{y}$ to denote points in $\mathbb R^d$.}
\end{quote}

\noindent 
There is a general and elegant theory of \emph{linear} PDEs, which is simpler in the absence of boundary conditions. For a linear (evolution) equation, $\frac{\partial u}{\partial t} = L u$, (i.e., first-order in time) of \emph{constant} coefficients, the solution takes the form
$$ u(\vec x, t) = \int_{\mathbb R^d} K(\vec x-\vec y, t) u_0(\vec y) \, d\vec y \,,$$
where $K$ is a function\footnote{known variously as the \emph{Green's function}, the \emph{propagator}, or the \emph{fundamental solution} of the linear operator $\frac{\partial }{\partial t} - L$. By definition, it satisfies $(\frac{\partial }{\partial t} - L)K=\delta$, where $\delta$ is the Dirac delta function, and $K$ may be thought of as the ``inverse'' to $\frac{\partial }{\partial t} - L$.}, or a \emph{generalized function} (Schwartz distribution), on $\mathbb{R}^{d+1}$, that depends only on the equation, not the initial data. The integral may be viewed as the convolution of $u_0$ with kernel $K(\cdot, t)$, for each time $t$. If $u$ is vector-valued, then $K$ is a \emph{matrix}-valued function, or a matrix of convolution operators. While for both the wave and the heat equations, the kernel $K$ can be written down explicitly, for any $d$ (see, for example, \cite{strichartz1994guide}), we shall only note the important difference: for the wave equation, $K(\vec x, t)$ is zero outside the cone\footnote{Such a region is known as the \emph{support} of the function or distribution, and corresponds to the so-called \emph{receptive field} in ConvNet.} $|\vec x|\leq |t|$, while for the heat equation, $K(\vec x, t)$ is always nonzero (only defined for $t>0$), although vanishingly small as $|\vec x|\to \infty$. In other words, the wave equation has \emph{finite speeds of propagation} in all directions, while the heat equation has an \emph{infinite} speed --- physically impossible, and should be taken as a mathematical idealization. The distinction gives rise to two separate ``classes'' of PDEs: the \emph{hyperbolic} and \emph{parabolic} equations, respectively.

Another characteristic difference is that, if the initial data is not smooth (infinitely differentiable), the heat equation will \emph{smooth out} any kinks or jumps immediately (see Fig.~2), while the wave equation preserves and spreads the discontinuities for all time. (This, too, can be ``read off'' from the kernel $K$.) A ConvNet is likely to involve both processes, even though we have emphasized the hyperbolic aspect in this paper.

The general formula also offers a (somewhat) different interpretation of the convolution operation in ConvNets that also applies to large kernels: instead of being a finite-difference approximation to \emph{differential} operators, the conv layers are \emph{integral} operators with these $K(\cdot, t)$ kernels, for discrete values of $t$. What is being learned is not so much the underlying equation, but the kernel $K$ that solves the equation.

If the coefficients in the operator $L$ are not constants, but \emph{variables} (i.e., functions of $\vec x$ and $t$), the equation is still \emph{linear} but the kernel $K$ would \emph{not} be a function of a single point $(\vec x, t) \in \mathbb{R}^{d+1}$, but of \emph{two points}: $ K \equiv K(\vec x, t, \vec y, s)$, which quite literally encodes how much the solution or the initial data at the point $(\vec y, s)$ \emph{contributes} to the solution at the point $(\vec x,t)$. In the case that $L$ is constant-coefficients, the kernel must then only depend on the difference $(\vec x - \vec y, t - s)$ by a simple argument with translation equivariance. 

It is, after all, not a coincidence that convolutions and kernels feature so prominently in the languages in both fields.

\subsection{Nonlinear PDEs}
Once we introduce nonlinearities into the equation or system, not only is there rarely a closed formula for the solution, there also does not exist anything close to a general theory, even just for hyperbolic type. Many aspects of the solution need to be considered on a case-by-case basis (not only on the equation, but also the function class to which the initial data belongs):

\begin{itemize}
\item[--] local existence (for a short time) 
\item[--] global existence (for all time) 
\item[--] uniqueness 
\item[--] regularity (smoothness) 
\item[--] continuous dependence on initial/boundary data 
\item[--] stability under small perturbations 
\item[--] \ldots
\end{itemize}

It is impossible to do justice to nonlinear PDEs in a few pages. One of the most celebrated phenomena, not confined to PDEs but shared with other nonlinear systems (of ODEs, or even discrete systems), is that the solution may be \emph{so} sensitive to the initial data that one can \emph{not} make accurate long-term predictions, even qualitatively. It has grown into whole fields of research, under the names of \emph{nonlinear dynamics}, \emph{chaos theory}, or \emph{complex systems}, though the parts that directly concern PDE are relatively nascent and less known. Naturally, one would like to bring the large body of work, both results and methods, to bear on the theory of neural networks, to have something useful to say, i.e., more than just importing terms such as ``emergence.''

On the opposite end of the spectrum, there is the (very rare) phenomenon of \emph{integrable systems}, exemplified by the KdV equation, that are characterized by exact solutions and rich algebraic structures. Would it be sensible to view the training of neural networks as finding the ``integrable'' ones within the large sea of chaotic systems?


\end{document}